\documentclass[sigconf]{acmart}

\usepackage{booktabs} 
\usepackage{multirow}
\usepackage{arydshln}

\setcopyright{acmcopyright}

\acmDOI{10.475/123_4}

\acmISBN{123-4567-24-567/08/06}

\acmConference[SIGSPATIAL'20]{ACM SIGSPATIAL conference}{November 2020}{Seattle, Washington USA} 
\acmYear{2020}
\copyrightyear{2020}

\acmPrice{15.00}

\begin{document}
\title{Geosocial Location Classification: Associating Type to Places Based on Geotagged Social-Media Posts}

\author{Elad Kravi}
\affiliation{%
  \institution{Amazon}
}
\email{ekravi@amazon.com}

\author{Yaron Kanza}
\affiliation{%
  \institution{AT\&T Labs-Research}
}
\email{kanza@research.att.com}

\author{Benny Kimelfeld}
\affiliation{%
  \institution{Technion}
}
\email{bennyk@cs.technion.ac.il}

\author{Roi Reichart}
\affiliation{%
  \institution{Technion}
}
\email{roiri@technion.ac.il}

\renewcommand{\shortauthors}{Kravi et al.}
\renewcommand{\shorttitle}{Geosocial Location Classification}

\begin{abstract}
Associating type to locations can be used to enrich maps and can serve a plethora of geospatial applications. An automatic method to do so could make the process less expensive in terms of human labor, and faster to react to changes. In this paper we study the problem of {\it Geosocial Location Classification}, where the type of a site, e.g., a building, is discovered based on social-media posts. Our goal is to correctly associate a set of messages posted in a small radius around a given location with the corresponding location type, e.g., school, church, restaurant or museum.  We explore two approaches to the problem: {\it (a)\/}~a pipeline approach, where each message is first classified, and then the location associated with the message set is inferred from the individual message labels; and {\it (b)\/}~a joint approach where the individual messages are simultaneously processed to yield the desired location type. We tested the two approaches over a dataset of geotagged tweets. Our results demonstrate the superiority of the joint approach. Moreover, we show that due to the unique structure of the problem, where weakly-related messages are jointly processed to yield a single final label, linear classifiers outperform deep neural network alternatives. 
\end{abstract}

%
%



\maketitle

\section{Introduction}
\label{sec:introduction}
Geospatial applications often use the type of geospatial entities to serve a request or compute an answer to a query, e.g., when searching for a nearby school there is a need to know which buildings are schools. The process of map enrichment adds information like type to objects on a map. A manual enrichment, however, can be labor-prone, slow and expensive, but it is not always easy to automatically classify sites and assign proper types to buildings and other geospatial entities. In this paper, we study the novel approach of assigning types to sites based on geotagged social-media posts.  

Over the last two decades, social media has become a prominent channel for information exchange and opinion communication. People share posts using microblog platforms like Twitter and Instagram about a variety of experiences and activities. The prevalence of smartphones has made such sharing widespread, so users share posts from diverse locations and at different times, and social-media platforms allow their users to tag posts with labels that indicate the location where each post has been created (\textit{geotagging}). The availability and timeliness of the information provided by geotagged posts attracted the attention of the research community, and this has resulted in growing popularity of tasks like information extraction from social-media posts~\cite{sakaki2010earthquake,xia2014citybeat}.

It has been shown that there is often a correlation between the location in which a post is created and the textual content of the post~\cite{10.1145/2619112.2619115}. Our goal is to use such correlations for automating the task of assigning a type to sites. This can be used in cases where a new map is created, when new objects are added to existing maps or when there are new types as part of a new classification.

Finding the type of locations, e.g., whether messages were posted from a restaurant, a museum or a church,  is a hard task. First, many messages are generic and do not relate to the geocontext. Second, messages are short and use an informal jargon. Third, messages by different users have different terminology and style. 

In this paper we study the task of {\it Geosocial Location Classification}. 
Given a set of geotagged messages sent from a certain location, our goal is to associate them with a location type from a pre-defined set of types. 
To explore this task we collected a dataset of tweets for a set of locations. Each location is defined as a certain radius around a given pair of coordinates, e.g., the center of a building. 
Our goal is to build a classifier that can learn the location type based on the content of the messages sent from each location and their context information (e.g., posting time.)

Our task provides a novel text classification challenge. While the classification of individual texts is widely explored (e.g., \cite{kim2014convolutional}), to the best of our knowledge we are the first to study the problem of learning location types from a set of weakly-related messages. 
 

Besides the known challenges associated with text processing of microblog messages (e.g., their short length, bad grammar and inclusion of emogies~\cite{Eisenstein2013}), our setup also has a unique challenge. Many messages published near a location are unrelated to the type of this location. For example, text messages posted from a school can be related to a broadcast of a sports event or to job-related issues of some parents. We relate to such messages as \textit{noisy messages} and aim to deal with this challenge through joint processing of multiple messages sent from the same location.
For example, detecting many messages written in a childish language can signal that the location is a school. In order to overcome this issue we analyze the properties of messages posted in different locations and design hand-crafted features that aim to capture location-level properties. 

We consider two approaches to our problem. 
In the \textit{pipline\/} approach, a classifier first classifies each individual tweet in the set and then an aggregation method infers the final location type of the entire set. In the \textit{joint\/} approach, a single classifier simultaneously processes all the messages in the set to infer the type of their location. For both approaches we implement methods based on traditional linear classifiers and methods based on state-of-the-art convolutional neural networks (CNNs), which have shown superior text classification performances in a number of studies (e.g.,~\cite{kim2014convolutional}). In addition, we examined four different sets of hand-crafted features---from surface textual features, through simple linguistic features to spatio-temporal features---as well as features based on word embeddings.

Our results 
demonstrate the power of the joint approach. Moreover, while CNNs and linear classifiers have similar accuracy in the classification of individual messages, we show that for Geosocial Location Classification, which requires the processing of a set of weakly-related messages, linear classifiers perform better than CNNs. Moreover, 
we show that the simplest surface and linguistic features are most useful for the task.

The main contributions of our work are as follows:
\begin{itemize}
\item Introducing and the problem of Geosocial Location Classification and its usages. 
\item Presenting a method to learn the class of a set of messages (tweets) and showing it outperforms baseline methods that initially classify a single message and then aggregating the set of labels.
\item Comparing standard and orthodox classifiers---re-justifying that CNN outperforms other methods for sentence classification, and showing that a logit classifier outperforms other classifiers for Geosocial Location  Classification. 
\end{itemize}

The paper is organized as follows. Section~\ref{section:related} surveys related work. Section~\ref{section:problem} defines the problem of Geosocial Location Classification and describes our methods. We elaborate on the learning models in Section~\ref{section:learning} and describe the dataset in Section~\ref{section:dataset}. Evaluation and results are presented in Section~\ref{section:evaluation} and conclusions in Section~\ref{section:conclusion}.
\section{Related Work}
\label{section:related}

In this paper we consider a specific type of the text classification task. The two properties we consider most crucial for our task are: (a)~the classification of a set of weakly related messages into a single class; and (b)~the association of a set of messages with a location type. In this section we survey the use of geottaged tweets, location classification, and text classification.

\medskip
\noindent
\textbf{Geosocial application.}
The use of geotagged social-media posts received a tremendous attention in recent years.
It has been shown how to use social-media posts for event and hot-spot detection~\cite{xie2013robust,zhu2016spatio,wei2018detecting,wei2019delle,becker2011beyond,sakaki2010earthquake,lee2010measuring,Benson:11}, measuring tourist activities in cities~\cite{kadar2014measuring,abbasi2015utilising}, finding recommended tourist attractions~\cite{alowibdi2014vacationfinder}, analyzing the dynamics of a city~\cite{cranshaw2012livehoods,kanza2014city,ferrari2011extracting,kling2012city}, keyword-based geographical search~\cite{pat2017s,magdy2014taghreed}, finding local news~\cite{10.1145/3139958.3141797,sankaranarayanan2009twitterstand}, location-based emotion analysis~\cite{doytsher2017emotion,sikder2019emotion,quercia2012tracking}, finding similarity between users based on their geotagged posts~\cite{10.1145/3054951,mizzaro2015content}, finding the home location of users~\cite{chang2012phillies,mahmud2014home,han2014text}, and so on. These many applications have shown how useful social media is, and in particular geotagged posts, for discovering new information about places, people and events.

\medskip
\noindent
\textbf{Geotagging tweets.} 
When geotagging social media posts the goal is to map posts without a geotag to the location in which they were created, based on the textual content. This has been studied in many papers~\cite{flatow2015accuracy,10.1145/3202662,Lieberman:10,Ahmed:13,10.1145/2065023.2065039,schulz2013multi},
see a recent survey in~\cite{zheng2018survey}. Many of these papers examine the language model of a place and compare it to the language model of a post, to associate the post with a location.
Geotagging posts, however, tries to detect particular locations and not types of locations, i.e., they try to find association of posts with a single place like Eiffel Tower while we try to associate a set of posts with a general type like ``school''. 


\medskip
\noindent
\textbf{Land-use classification.}
Some papers studied the use of geotagged photos for classification.
Hu et al.~\cite{hu2015extracting} studied the discovery of areas of interest in an urban environment, based on geotagged photos. 
Land-use classification based on geotagged photos was studied in~\cite{10.1145/2390790.2390794}. In all these cases, the algorithms rely on images rather than text. Note that using images does not work well for classifying buildings that look similar, especially in an urban area.

\medskip
\noindent
\textbf{Text classification: Features and models.} 
Text classification has been extensively studied in the NLP literature for many years. The standard approach to this task has long been to manually extract features from the text and use this representation for classifier training~\cite{manning2003optimization}. Previous works on the task differ in the features they used and the classifiers they employed.

The most common representation in text classification is bag-of-words. In recent years this approach was extended in various ways. For example, it was augmented with sparsity-inducing regularizers~\cite{Yogatama:14a} and information about (latent) linguistics structures such as parse trees, topics, and hierarchical word clusters~\cite{Yogatama:14b}. Surveying the various methods based on this representation is beyond the scope of this paper. In our models when using hand-crafted features we consider the vanilla bag-of-words features, as well as features based on POS tags and on language modeling.

Recently, NNs were proven effective in solving text classification tasks. Some methods learn word embeddings~\cite{Mikolov:13,pennington2014glove} and derive a text representation that is then employed in a classifier~\cite{Arora:17}. Other methods directly learn a vector representation of the entire text~\cite{dai2015document,kiros2015skip,tai2015improved} and feed it to a classifier. Such methods sometimes include direct application of recurrent neural networks (RNNs and LSTMs~\cite{hochreiter1997long})~\cite{kiros2015skip,Ziser:18} and auto-encoders~\cite{Glorot:11,Ziser:17}. 



A particular challenge our task poses for structure-aware NNs such as CNNs and LSTMs is that the messages we consider for each location are only weakly-dependent, as they are authored by different people and at different times.  Our experiments with CNNs, which demonstrated state-of-the-art results in many types of text classification tasks, when compared both to classifiers with hand-crafted features and to other types of NNs~\cite{zhang2015character}, demonstrate that our task is still challenging for structure-aware NNs.

\section{Overview}
\label{section:problem}

In this section we present the setting. We provide notations and define the research problem.

\subsection{Problem Definition} 
Let $L$ be a given set of location types, for example, 
$$L=\{\textit{restaurant}, \textit{school}, \textit{museum}, \textit{church}\}.$$ 
We also refer to types as labels. Let $E$ be a set of geospatial entities (sites/locations/places). We assume that each entity has a point location, e.g., for a building we use its center of mass. We denote by $\lambda(e)$ the location of entity $e\in E$. The goal of classification is to create a mapping from the geospatial entities in $E$ to labels in $L$. 

In this paper, the classification is based on geotagged social-media messages.
A geotagged message is denoted by $m = (l,t,c)$, where $l$ and $t$ are the location and time at which the message was created, in correspondence, and $c$ is its textual content. We denote by $\lambda$ the function that assigns a geotagged message to its location, i.e., $\lambda(m)=l$. Given a set $M$ of geotagged messages, the messages associated with an entity $e\in E$ are the messages whose distance from $E$ does not exceed a given threshold $r$, that is, $M(e)=\{m\in M \mid \left\|\lambda(m) - \lambda(e)\right\|\leq r \}$.

A {\em ground truth\/} is a given mapping $g:E\rightarrow L$ from entities in $E$ to labels in $L$. A ground truth can be based on existing labeled map objects or on labeling by human experts.
The ground truth is only available for a small set of entities and is used for {\em training\/} and {\em verifying\/} a classifier. In the training phase, the goal is to take entities, the messages associated with them and the label assigned to them in the ground truth, and based on that train a classifier that maps other entities to labels in $L$. In the verification phase, the accuracy of the classifier is measured by comparing the classification results to the ground truth. 

In the verification, a correct mapping is when the type assigned by the classifier is equal to the type according to the ground truth. For a given set of entities with a ground truth, the per-class {\em precision\/} is the percentage of correct mappings among the locations associated with the class. The per-class {\em recall\/} is the percentage of correct mappings among the locations that truly belong to the class. The $F_1$ score is the harmonic mean of recall and precision. In the learning phase, the aim is to build a classifier that would yield classification with high recall, precision and $F_1$ score in the verification. We apply a machine learning approach to this problem, and set the goal to learn a classifier $C:(e,M(e))\to L$. We use a standard supervised approach for minimizing the loss for a labeled training set. 


\subsection{Learning Approaches}

\begin{figure}[t]
  \centering
  \includegraphics[width=0.45\textwidth]{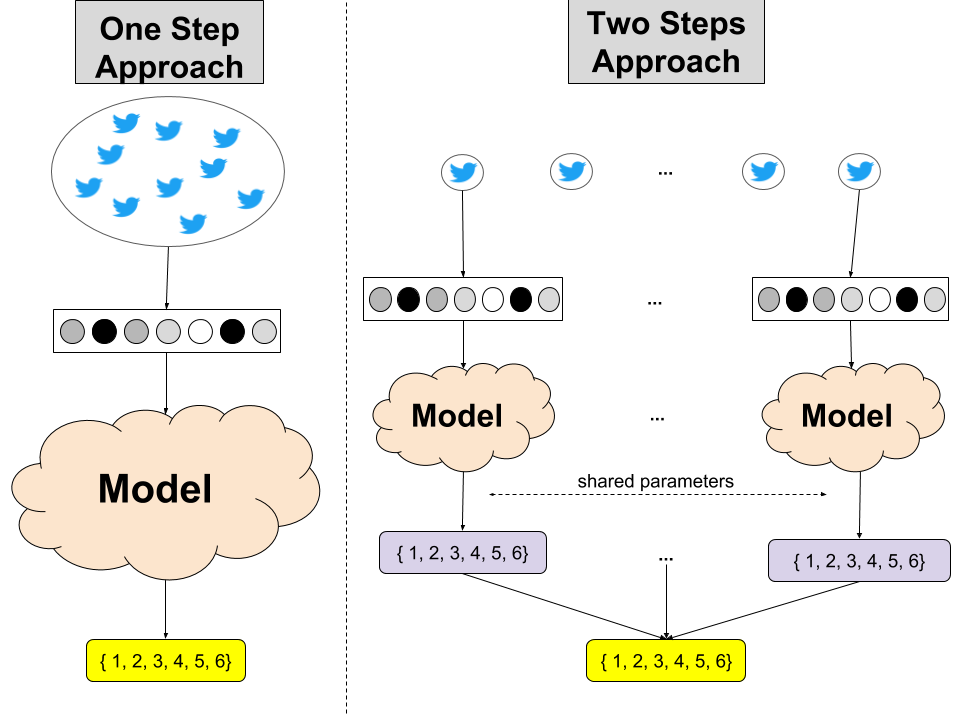}
  \caption{\label{fig:approaches} One vs. Two Step Classification.} 
\end{figure}

As mention in Section~\ref{sec:introduction}, the two main challenges in Geosocial Location Classification are as follows. First, coping with noisy messages within $M$, where the noise is defined with respect to the task. Noisy messages are messages that are sent from the proximity of a location, but their content is not related to the location type.  
Second, learning associations among messages in $M$ that could assist classification, e.g., seeing that a large portion of the messages posted from a site on Sundays may provide a supporting evidence that the place is a church. 

We now describe two approaches to tackle these challenges. The two approaches are depicted in Figure~\ref{fig:approaches}.
(I)~\emph{Pipeline) Classification}: Two step classification that consists of classifying each message in the set $M$ and then classifying $M$ based on the classifications of its elements; and (II)~\emph{Joint Classification}: A single step classification by simultaneously processing all the messages in the set $M$ and deciding on the location type of $M$.

\medskip
\noindent
\textbf{Pipeline Classification.} In pipeline classification we partition the problem into two easier problems. First, we classify each single message by training a message-level classifier, and then, classify the set $M$ based on the labels that were assigned in the first step. In the second step we compare the following two distributions: (i)~the predicted label distribution according to the classifier of the first step, and (ii)~to the label distribution of each location type in the training set. See details in Section~\ref{section:learning}. 

The pros and cons of this approach stem from its simplicity. Particularly, it decomposes the processing of the content of individual messages (first step) from the global decision that takes into account the label distribution of different location types (second step). While this facilitates two simple classification steps, it also results in partially-informed decisions at each step.



\medskip
\noindent
\textbf{Joint Classification.} In joint classification we learn a single model that jointly uses both the content of individual messages and associations between messages. The pros and cons of this approach are complementary to those of the pipeline approach. Particularly, when jointly considering the entire set of messages, it is hard to set any particular order between the messages: messages posted in the same location are authored by different users and their temporal order does not necessarily reflect the connection between them. While the Joint Model is more complex than the Pipeline Model, it does consider the content of the various messages in $M$ when making the location-type decision and does not decompose the problem into a series of local problems as the Pipeline Model does.

The difficulty in implementing the Joint Model is that the number of messages varies in different sets, that is, different locations are associated with a different number of messages. We considered two methods for implementing the Joint Model.
\begin{itemize}
\item \emph{Messages Concatenation:} The messages from a single location are concatenate to create one long {\em multi-message}. A model is trained over multi-messages and it classifies multi-messages, that is, the model discovers the location type of given multi-messages. Note that concatenating messages creates an order between the messages, however, this order is arbitrary and may affect the results arbitrarily.
\item \emph{Embed whole messages:} Instead of embedding single words, embed whole messages. The model is trained over sets of embedded messages from a single location and it classifies the location, i.e., assigns a label to the location. An approach for embedding whole sentences was proposed by Dai et al.~\cite{dai2015document}.
\end{itemize}





\section{Models}
\label{section:learning}

In the previous section we described the problem of Geosocial Location Classification, where the input is a set of geotagged messages and the goal is to correctly label this set. 
We now describe the models that are trained to solve this classification task, using one-step and two-step classification. Previous studies have shown Convolution Neural Networks (CNN) can outperform other classification models in the task of sentence classification~\cite{zhang2015character}. We evaluated this for the task of location classification by comparing orthodox classifiers to various types of neural networks. Furthermore, we designed an ad hoc model for the one-step classification, for both orthodox classification and neural networks.

In this section we describe our models. 
We design our models so that we can answer three related questions. (1)~Which approach performs better: Pipeline Model or Joint Model? (2)~Which class of models is more suitable, linear classifiers or deep neural networks (DNNs)? (3)~Which features are most suitable for our task? The first two questions are tightly connected since a major challenging property of our task is that the messages from a given location do not conform to an easily identified structure (see Section~\ref{section:problem}). 

\subsection{Linear Classification}
\label{section:linear}

We first describe our feature-set and we will provide more analysis in Section~\ref{section:dataset}. We then discuss the implementation of the Pipeline Model and the Joint Model.

\medskip
\noindent
\textbf{Feature sets.} We consider four feature sets. The set of \textit{basic textual features\/} consists of surface-level features such as the number of words and the number of characters in the message (binned into 0-10, 11-20, 21-30 and 31+), as well as counts of non-alphanumeric characters. The set of \textit{n-gram features\/} consists of counts of word n-grams, where we considered $n=1,2$, with a threshold of 5 appearances in the training set. 

The set of \textit{NLP features} consists of language model scores and part of speech tags.
A language model (LM) is a distribution over sequences of words in the corpus. Different distributions are associated with different types of entities. For example, in a school we expect many uses of words like `teacher', `class', `exam', `grade', etc. In a healthcare facility we expect to see words like `doctor', `nurse', `sick', `physical exam', etc.
In other words, given a corpus $\mathcal{D}$ comprising a set of documents, a language model $M_\mathcal{D}$ represents the probability of the corpus to generate a given text. The model assign a probability proportional to the occurrences of terms in $\mathcal{D}$~\cite{manning2008introduction}. $KL$-divergence is a common metric to capture the differences between language models of different classes~\cite{kullback1951information}. Given two discrete probability distributions $P_1$ and $P_2$ defined on the same probability space $\mathcal{S}$, the  the Kullback-Leibler divergence from $P_1$ to $P_2$ is defined as 
$$D_{\textit{KL}}(P_2|P_1) = \sum_{s \in \mathcal{S}}P_2(s)\log \left(\frac{P_2(s)}{P_1(s)}\right)$$ 
 
Language model scores are computed for each location type $\mathcal{T}$ by training an LM $M_\mathcal{T}$ on the messages associated with the location in the training set. Then, we add a score to each message $m$ in the training set according to how similar it is to each class-specific LM. Scores are calculated using the query likelihood (QL) metric~\cite{ponte1998language}, 
$$QL(m,M_{\mathcal{T}})=Pr(m|M_{\mathcal{T}})=\prod\limits_{t\in m}{p(t|M_{\mathcal{T}})},$$ 
where $t$ are the tokens of $m$, $p(t|M_{\mathcal{T}})$ is the probability of term $t$ under the language model of class $\mathcal{T}$,
and the term probabilities are computed with Dirichlet smoothing~\cite{manning2008introduction}. For using Parts-of-Speech (POS) tags, we counts for each message the number of appearances of each one of the following $13$ major POS tags: ``CD''  (cardinal numbers), ``DT''  (determiners), ``FW'' (foreign words), ``IN'' (prepositions and subordinating conjunctions), ``JJ''  (adjectives), ``NN''  (common nouns), ``NNP''  (proper nouns), ``NNS'' (common noun plural form), ``PRP'' (personal pronoun), ``RB'' (adverbs), ``VB''  (verb base form), ``VBG''  (verb gerund or present participles), ``VBP'' (verb non-3rd person singular present form). 

The set of \textit{spatio-temporal features\/} uses distance from the classified entity, where the Haversine formula~\cite{korn2000mathematical}
is applied in order to compute the distance between messages and classified entities.
For the temporal feature, we use the time of day in which the message was posted.
We tested in our experiments each combination of the feature sets, and report results with the combination that performs best on the development set.


\begin{table}[t]
  \renewcommand{\arraystretch}{1.2}
  \begin{tabular}
  {|c|c|c|}
  \hline
   Family name & Description & \# features \\
   \hline\hline
   textual & standard textual features & $32$ \\
   $n$-grams & uni-grams and bi-grams & $\sim 50000$\\
   LMs & language models scores & $6$ \\
   POS & part-of-speech tags & $13$ \\
   spatio-temporal & distance from entity, day period & 2 \\
  \hline
  \end{tabular}
  \caption{\label{table:features} Feature families.}
\end{table}

\medskip
\noindent
\textbf{Pipeline classification.} 
To implement the pipeline classifier,
in the first step we train a classifier for individual messages using the above features. We examined the use of both Logit~\cite{hausman1984specification} and Naive Bayes~\cite{murphy2006naive} classifiers. When executing these classifiers, we assign to each message the label with the maximal probability according to the model. (Code for the classifiers is taken from \url{https://nlp.stanford.edu/wiki/Software/Classifier}.)

In the second step, for each location there is the multiset of labels created in the first step. The proportion between the labels is used to decide the overall type. Note that because of noise, not all the messages in a location are related to the type of the location. For example, in a school we  expect to see a certain percentage of messages related to school, a different percentage of messages related to shopping, a certain percentage of messages related to healthcare issues, etc. In a healthcare facility we expect these percentages to be different from those received in a school, i.e., different proportions between the sets associated with the different location types. In a Pipeline Classifier, we learn the expected percentages and compare the observed percentages per each site to the expected ones, to decide what is the type of the site. We select as the overall label per location the label with highest positive difference between the observed and the learned proportions.


\medskip
\noindent
\textbf{Joint classification.} In Joint Classification, we first concatenate all the messages associated with the location and then compute the n-gram features, LM-based and POS-based features. 
We then train the Logit and Naive Bayes classifiers.



\subsection{Deep Neural Networks}
\label{section:neural}

The linear classifiers reflect a structure-ignorant approach. That is, since it is hard to characterize the structure of the message set, linear classifiers ignore relationships between messages and treat the messages as independent of each other. To examine a method that does take connections between messages into account, we tested classification using deep neural networks (DNNs).

We compare two types of DNNs---convolution neural networks (CNN)~\cite{lecun1998gradient} and recurrent neural networks (RNN), including LSTM and bi-directional LSTM~\cite{chiu2015named}. While we explored a variety of recurrent network architectures, they were all substantially outperformed by the CNNs. We hence do not report results for this class of methods.

We now describe the architecture of the networks.
The input for the CNN is a tensor of word embeddings. We used Glove embeddings (\url{http://nlp.stanford.edu/data/glove.twitter.27B.zip})~\cite{pennington2014glove}, trained on a Twitter dataset with dimension of $200$.
Following~\cite{kim2014convolutional}, we employ $K$ filters with sizes of $d={3,4,5}$ times the embedding domain. Each filter is iterating over the input matrix using a sliding window while generating a $1 \times (n-d+1)$ size vector, where $n$ is the input text length. A max pooling is then performed for each of the $K$ vectors, to generate a single $1 \times K$ vector that is fed into the classification (softmax) layer. Our DNN models were implemented with Deeplearning4j (see~\url{http://deeplearning4j.org}).


\medskip
\noindent
\textbf{Pipeline classification.} We store the embeddings of the words of each message in a matrix of size $|W|\times d$ where $|W|$ is the number of words and $d$ is the embedding dimension. After classifying each message we apply the same distribution-based approach described for pipeline classification in Section~\ref{section:linear}.

\medskip
\noindent
\textbf{Joint classification.} Following~\cite{dai2015document} we averaged the embeddings of the words in each message.  For comparison, we also implemented this approach where each tweet is represented by the sentence embedding method of~\cite{inferSent:17} (see \url{https://github.com/facebookresearch/InferSent}) with the same word embeddings fed to the sentence embedding algorithm. The columns of the resulting matrix contain the embeddings of all the messages sent from a single location.

Like with the Linear Classifiers, the pipeline approach again reflects a structure-ignorant approach.  The joint classification approach, on the other hand, assumes a connection between the messages. As pointed by \cite{DBLP:journals/corr/abs-1802-05187} the input order for a CNN impacts its  performance, and in our case we expect a particular importance for this order as it reflects the relations between the messages sent from the classified location. 

We considered three message sorting methods. (1)~Sorting by distance from the classified location, starting from nearest to the most distant message.
(2)~Sorting by LM score. To that end, we compute the scores given by the LMs of all location types and associate a message with the LM that is most similar to it. Then, we order the messages according to the location types they were assigned to, in decreasing order of LM scores within each class. The order of the per-location message sets in the input matrix was randomly set. (3)~Using a random order.

Interestingly, in our experiments we observed that all three sorting methods had similar accuracy. Hence, we report results for the most parsimonious random ordering approach, and mark the message ordering problem as an important open challenge.


\section{Dataset}
\label{section:dataset}
This section describes the dataset we used for training the model and for the experiments. It provides background information to the learning models explained in Section~\ref{section:learning}.

\begin{figure}[t]
  \centering
  \includegraphics[width=0.45\textwidth]{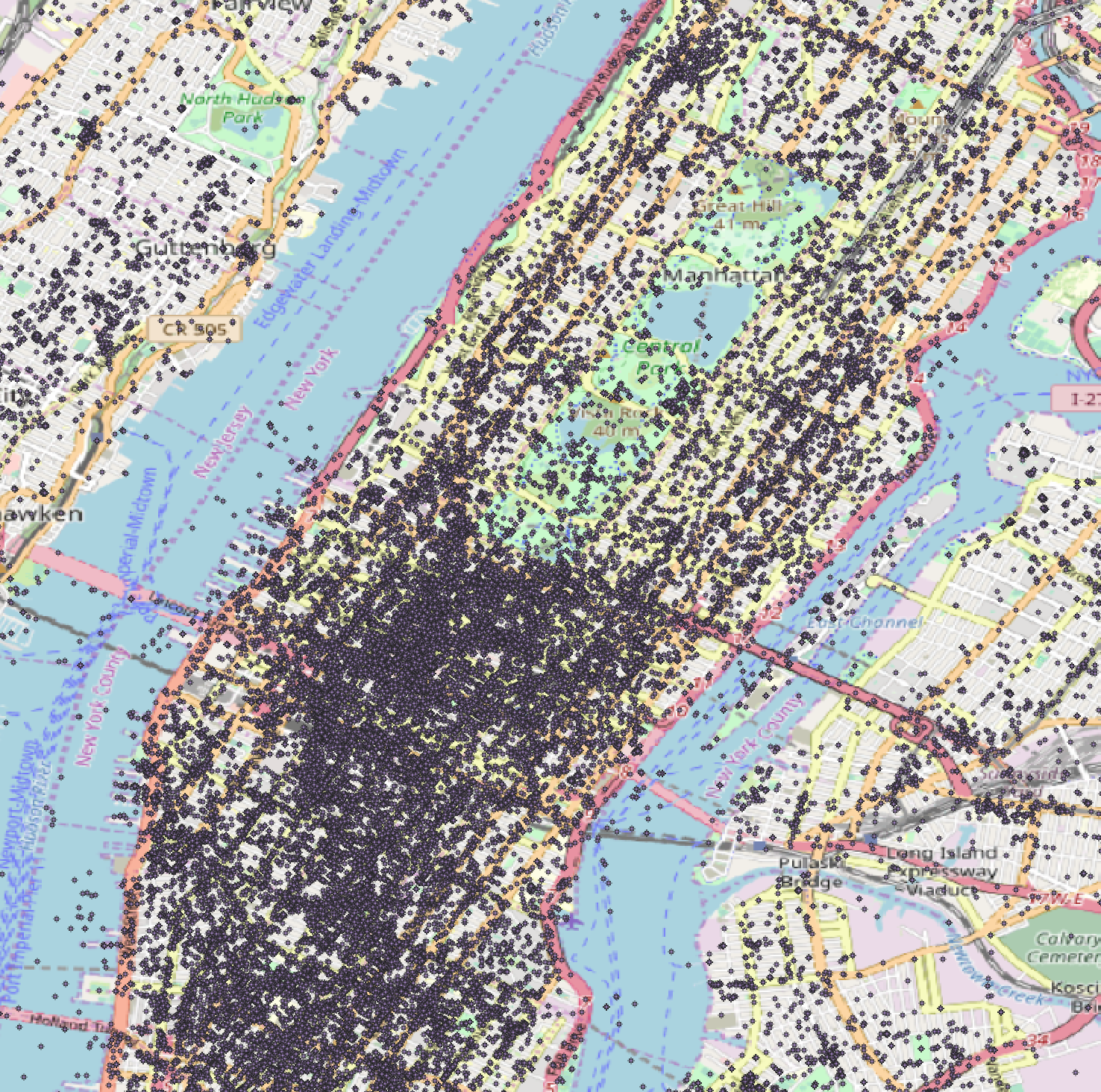}
  \caption{\label{fig:messages} Geospatial distribution of the geotagged tweets.} 
\end{figure}

\begin{table*}[t]
\centering
  \renewcommand{\arraystretch}{1.2}
  \begin{tabular}
  {|c|c|c|c|c|c|c|c|c|}
  \hline
   & & schools & universities & churches & shops & museums & health & total\\
  \hline
  \multirow{ 2}{*}{locations} & \# & 323 & 79 & 152 & 97 & 64 & 273 & 988\\ 
    & \% & 32.7 & 8 & 15.4 & 9.8 & 6.5 & 27.6 & - \\
    \hline
    \multirow{ 2}{*}{tweets} & \# & 9626 & 7811 & 37088 & 44038 & 27560 & 11612 & 137735\\ 
    & \% & 7 & 5.7 & 26.9 & 32 & 20 & 8.4 & - \\
    \hline
    $\left\lceil\frac{\text{\#tweets}}{\text{location}}\right\rceil$ & & 59 & 174 & 414 & 1142 & 1276 & 92 & 139\\ 
\hline 
  \end{tabular}
\caption{ \label{table:dataset} The distribution of locations, tweets and type-to-token ratio for the tweets of each location type.}
\end{table*}
 
Our dataset consists of messages posted on Twitter (\url{www.twitter.com}). The dataset consists of $14.5$ million geotagged tweets from the Manhattan area, collected using the developers API during a period of $400$ days, in the years 2013 and 2014.
See the spatial distribution of the tweets in Figure~\ref{fig:messages}. Each message record consists of textual content, coordinates from which it was sent and a time-stamp.

In our analysis, we consider six location types: schools, universities, churches, health locations, large shops and museums. These types were chosen since they 
are public places where the activity is expected to be high and distinguishable. That is, frequent terms used in a hospital are expected to be different from those used in a museum.
The location of each entity was given by 
the NYC Open Data website (\url{https://opendata.cityofnewyork.us}) and by the Data Gov website (\url{https://www.data.gov/}). The classified locations are the center-of-mass points of the entities. 

We collected all the tweets within a distance of $20$ meters around each location. This distance was chosen to address the  typical horizontal error of positioning using civilian Global Positioning Systems ($5-15$ meter, according to \url{https://www.gps.gov/systems/gps/performance/accuracy/}), while maximizing the association between messages and the location they were sent from. Locations associated with less than $5$ tweets were not included. 

We observed a large variance in the number of messages posted from locations of the same type. For example, shops located near Times Square are typically associated with many more tweets than shops located in the upper west side. Hence, we applied the following process: let $|M_v|$ denote the number of messages posted in a location $v$ and let $|\overline{M_{L_v}}|$ denote the average number of tweets for $v$'s location type. If $|M_v|$ was smaller than $|\overline{M_{L_v}}|$ we included all the messages $M_v$ in the dataset; otherwise, we picked  $|\overline{M_{L_v}}|$ random messages from $M_v$.
The resulting dataset consists of $988$ locations and about $138000$ tweets (see Table~\ref{table:dataset}). 




Next, we analyze properties of our dataset that might be useful for characterizing the differences between location types. This will  help us select classification features.

\medskip
\noindent
\textbf{Language modeling.} In order to characterize the unique lexical properties of each class (location type) we performed a language model analysis, to find the distribution over sequences of words in the corpus. Particularly, we train bi-gram LMs. We used our own implementation of the LM, which includes Laplasian smoothing and backoff to unigrams for each location type. For a classified type, we compare the computed LM to the LM that is trained on all the messages in our training set. We then consider the $k$ terms that contribute the most to the $KL$-divergence~\cite{kullback1951information} between the two types of LMs. The results are presented in Table~\ref{table:klDivergence}.


It can be seen that some bi-grams represent the venue type (``High School'', ``York Academy'', ``New Museum'', ``Cathedral of'', ``Community Health'') while others represent a specific venue (``Cornell Medical'', ``Apple Store''). 
However, many other bi-grams are seemingly irrelevant to our task (``Below New'', ``86th Street'', ``Central Terminal'', ``City of'', ``Junior League''), and were probably generated by noisy tweets (as mention in Section~\ref{sec:introduction}). Hence, lexical information seems to be relevant but not sufficient for our task.

\begin{table*}[t]
 \centering
\small
\renewcommand{\arraystretch}{1.2}
\begin{tabular}
  {|l|l|l|l|l|l|}
\hline
   \textbf{Schools} & \textbf{Universities} & \textbf{Churches} & \textbf{Shops} & \textbf{Museums} & \textbf{Health}\\
  \hline
   Manhattan/Upper East & Cornell Medical & Church of & Apple Store & Museum of & Cinemas 86th\\ 
   of St. & York Institute & of St. & @ Manhattan & Intrepid Sea, & Bard Athletic\\ 
   High School & NewYork-Presbyterian/Weill Cornell & John the & Grand Central & Sea, Air & Community Health\\ 
   Road Runners & York Academy & St. John & Manhattan, NY & Air \& & at Bard\\ 
   Riverside Church & Assembly West & Memorial Preview & at Apple & \& Space & Athletic Center\\ 
   York Road & New Work & Preview Site & Forbidden Planet & New Museum & Callen-Lorde Community\\ 
   Divine, NYC & Work City & @ Cathedral & Central Terminal & Space Museum & 86th Street\\ 
   Runners - & Medical Center & the Divine & Dover Street & @ Intrepid & York Junior\\ 
   the Divine, & Academy of & Cathedral of & Street Market & City of & Junior League\\ 
   Luke in & Below (New & Roman Catholic & Manhattan, New & @ Museum & Harlem World\\ 
  \hline
   \end{tabular}
     \caption{\label{table:klDivergence} Most distinguishing bi-grams based on the $KL$-divergence between the class-specific LM and the corpus LM.}
\end{table*}

\medskip
\noindent
\textbf{Distance from entity location.} We next consider the distance of each message from its location. Our assumption is that distances may be indicators of relevance and hence the closer the messages of a given location are to its coordinates, the stronger the signal that we can expect from the content of these messages. We aggregated the tweets to three classes: \emph{adjacent\/} with distance of up to $5$ meters, \emph{near\/} with distance of $5-12$ meters and \emph{far\/} with distance of $12-20$ meters from the classified location. See results in Table~\ref{table:distances}.

We found that the majority of the tweets in museums were sent from adjacent locations ($49\%$), while the majority of tweets from schools, universities and health facilities were sent from far locations ($65.4\%, 66.2\%$ and $65.6\%$, respectively). The distances of tweets sent from shops and churches are roughly evenly distributed between the classes. 

We hypothesize that the reason for this pattern is that museums tend to be isolated, while educational and health institutes may restrict usage of cellphones within nearby surroundings. This may also indicate the coherence of tweets sent from museums in comparison to the latter set of location types as is reflected in the LM analysis of Table~\ref{table:klDivergence}.

 \begin{table}[t]
 \centering
   \renewcommand{\arraystretch}{1.2}
   \begin{tabular}
   {|l|c|c|c|}
   \hline
    & adjacent ($<$5m) & near (5--12m) & far (12--20m)\\
   \hline
 schools	& $6.59\%$ & $27.99\%$ & $\boldsymbol{65.42\%}$\\
 universities & $7.28\%$ & $26.48\%$ & $\boldsymbol{66.24\%}$\\
 churches & $31.55\%$ & $24.74\%$ & $43.72\%$\\
 shops & $26.86\%$ & $24.30\%$ & $48.84\%$\\
 museums & $\boldsymbol{49.32\%}$ & $18.66\%$ & $32.02\%$\\
 health & $5.44\%$ & $28.93\%$ & $\boldsymbol{65.64\%}$\\
 \hline 
   \end{tabular}
   \caption{\label{table:distances} Aggregated distance from entity.}
 \end{table}

\begin{figure}[t]
  \centering
	\fbox{
  \includegraphics[width=0.45\textwidth]{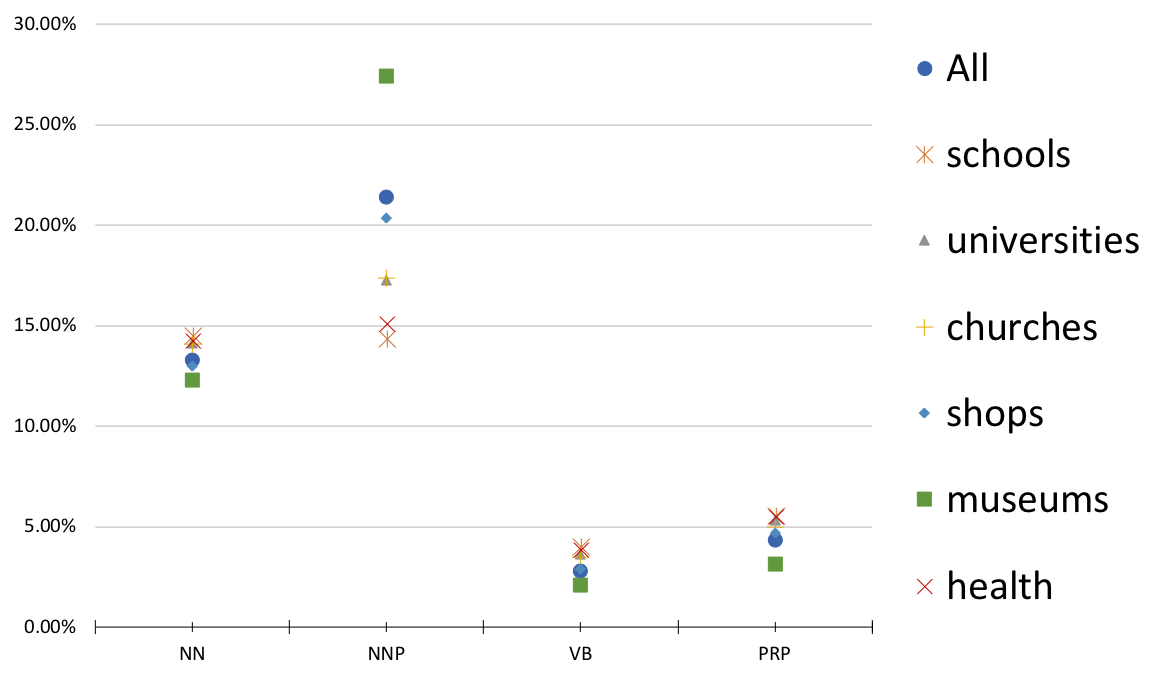}
	}
  \caption{\label{fig:pos-tags} POS prevalence per location type.} 
\end{figure}

\medskip
\noindent
\textbf{Part-Of-Speech distribution.}
We applied the Stanford's POS tagger~\cite{toutanova2003feature} to messages sent from each location type. Figure~\ref{fig:pos-tags} compares the prevalence of four major POS tags (NN, NNP, VB and PRP) in the different location types. We also measured other POS tags like adjectives, but no significant difference was found between the classes.

POS tags prevalence in different location types is presented in Table~\ref{table:pos}. It can be seen that messages sent from schools have a larger proportion of (non-proper) nouns, verbs and pronouns and a smaller proportion of proper nouns, indicating a simpler language (e.g., ``I need to get my life together right about now lol'', ``Dear iphone, I am never trying to say ducked up'') which  may be due to the young age of the content creators. This finding is in line with previous studies about language learning~\cite{collins2004language}. Universities and churches have similar POS distributions. Museums have a large proportion of proper nouns and a smaller proportion of personal pronouns in comparison to the overall statistics ($27.4\%$ versus $21.4\%$ and $3.1\%$ versus $4.3\%$). This can be explained by the many names associated with art (e.g., famous painters and artists). Health locations have a smaller proportion of proper nouns, and a larger proportion of verbs. Messages sent from shops have a similar distribution to that of the entire corpus, indicating a general, non-unique language.

 \begin{table*}[t]
 \centering
   \small
   \renewcommand{\arraystretch}{1.2}
   \begin{tabular}
   {|c|c|c|c|c|c|c|c|}
   \hline
    POS tag & all & schools & universities & churches & shops & museums & health \\
   \hline
 	NN & 13.29\% & \textbf{14.54\%} & 14.12\% & 14.07\% & 13.02\% & 12.29\% & 14.27\% \\
 	NNP & 21.42\% & \textbf{14.38\%} & 17.29\% & 17.39\% & 20.34\% & \textbf{27.41\%} & \textbf{15.12\%} \\
     VB & 2.77\% & \textbf{4.02\%} & \textbf{3.64\%} & \textbf{3.51\%} & 2.88\% & 2.08\% & \textbf{3.89\%} \\
     PRP & 4.33\% & 5.57\% & 5.33\% & 4.96\% & 4.70\% & \textbf{3.14\%} & 5.54\% \\
 \hline 
   \end{tabular}
   \caption{\label{table:pos} POS tags prevalence in different venue types comparing to prevalence in all the messages.}
 \end{table*}

\begin{figure}[t]
  \centering
	\fbox{
  \includegraphics[width=0.45\textwidth]{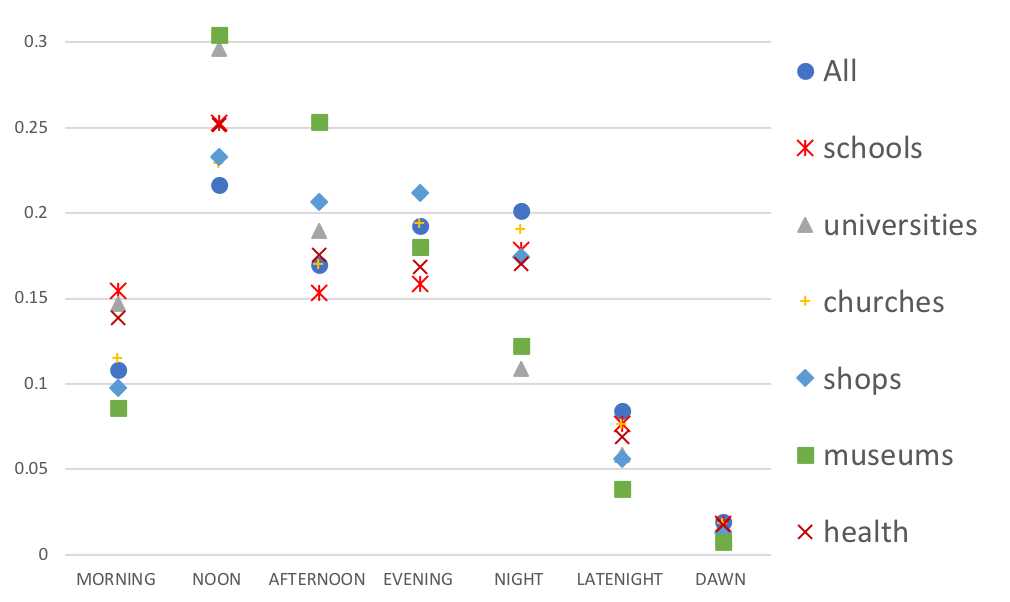}
	}
  \caption{\label{fig:day-period} Temporal analysis showing the portion of messages as a function of sending time, for each location type.} 
\end{figure}

\medskip
\noindent
\textbf{Temporal analysis.}
We further categorized the tweets to six categories, reflecting their sending time.     
\emph{Morning\/} relates to 7--11 AM, \emph{Noon\/} refers to the time between $11$ AM and $3$ PM, \emph{Afternoon\/} is 3--6 PM, \emph{Evening\/} refers to 6--9 PM, \emph{Night\/} is $9$ PM to midnight, \emph{Late night\/} is midnight to $4$ AM and \emph{Dawn\/} is associated with 4--7 AM.
Figure~\ref{fig:day-period} presents the class distribution of the messages of each location type and of the entire set of messages.

As expected, schools and universities are more active during morning and noon time, museums are more active during noons and afternoons and shops are more active during afternoon and evening times. Yet, while the temporal signal is valuable and used as a feature as described in Section~\ref{section:linear}, it cannot differentiate the classes without further information from other sources. 

\section{Evaluation}
\label{section:evaluation}

\begin{table}[t]
\centering 
  \renewcommand{\arraystretch}{1}
  \begin{tabular}
  {|l|l|c|c|}
  \hline
   Model & Configuration & Accuracy(\%) & $F_1$ \\
  \hline
  majority  & (majority class is school) & 32.7 & 8.2 \\
  random & & 16.7 & 16.7 \\
	\hdashline
  logit & textual \& $n$-grams & 51.3 & 48.5 \\
  logit & $n$-grams & \textbf{52.6} & \textbf{48.8} \\
  Naive Bayes & textual \& $n$-grams & 27.0 & 18.3 \\
	\hdashline
  CNN &  word embd.& 36.4 & 20.7 \\	
  CNN & sentence embd. & 32.8 & 36.0 \\
\hline 
  \end{tabular}
	\caption{\label{table:result:1}Joint Classification.} 
	\end{table}
  
\begin{table}[t]
\centering 
  \renewcommand{\arraystretch}{1}
	\begin{tabular}
  {|l|l|c|c|}
  \hline
   Model & Configuration & Accuracy(\%) & $F_1$ \\
  \hline
  majority  & (majority class is school) & 32.7 & 8.2  \\
  random & & 16.7 & 16.7  \\
	\hdashline
Naive Bayes & textual \& $n$-grams & 23.3 & 28.7  \\
  Naive Bayes & $n$-grams \& spatio-textual & \textbf{27.7} & \textbf{34.7}  \\
  logit & $n$-grams \& spatio-textual & 13.6 & 22.8  \\
	\hdashline
   CNN & word emb. & 26.2 & 26.8\\
\hline
  \end{tabular}
	\caption{\label{table:result:2}Pipeline Classification. }
	\end{table}
  
\begin{table}[t]
\centering 
  \renewcommand{\arraystretch}{1}	
	\begin{tabular}
  {|l|l|c|c|}
  \hline
   Model & Configuration & Accuracy(\%) & $F_1$ \\
  \hline
  majority & (majority class is shop) & 32.0 & 8.1 \\
  random & & 16.7 & 16.7 \\
  \hdashline
	logit & textual \& $n$-grams & 55.0 & \textbf{45.0} \\
  Naive Bayes & & 47.4 & 42.0 \\
  \hdashline
	CNN & word embd. & \textbf{55.5} & 43.5 \\
\hline 
  \end{tabular}
  \caption{\label{table:results} Classification of individual messages.} 
\end{table}

We present now the results of the tests we conducted.
The goals are to compare the different methods and examine the accuracy of these methods for Geosoacial Location Classification.
We report results with two measures, accuracy and class-based $F_1$-score. We run a 10-fold cross-validation protocol with random sampling across locations for Joint Models and across messages for Pipeline Models, and report the averaged result of each model across the 10 folds. The train/dev/test ratio in each fold is 64:16:20. 

As a baseline for comparison, we use the {\it majority\/} and {\it random\/} classifiers.
The {\it random\/} classifier randomly selects the location type per each location. The majority classifier maps all the locations to the location type with the largest number of instances. For example, suppose that stores are the most common entities, that is, have more instances than other location types. A majority classifier would classify every element as a shop. Clearly, a useful classifier must be much better than the random and majority classifiers.  

Table~\ref{table:result:1} presents classification results for Joint Classification. The majority class is school and we can see that a naive classifier based on majority achieves more that 32\% accuracy. A logit classifier over n-grams provides the best results in this case.  

In Table~\ref{table:result:2} we see the classification results for Pipeline Classification. We can see that the classification accuracy is highest when using a Naive Bayes Classifier over n-grams and the spatio-textual features. Overall, the classification results for the Pipeline Model are much lower than those of the Joint Model. 

In Table~\ref{table:results} we see the results of classification of individual messages, e.g., map a message that is posted from a school to the location type `school'. We can see that in this task, CNN using word embedding provides the highest accuracy, slightly better than a logit classifier over the textual features and the n-grams.

Tables~\ref{table:result:1}-\ref{table:results} present the classification results for the different classifiers and shed light on the questions we raised throughout the paper. First, Joint Classification (one-step) substantially outperforms the Pipeline Model, with the best Joint Model (logit with n-grams and textual features) achieving accuracy and $F_1$-score of 52.6 and 48.8, respectively, while the best Pipeline Model (two-steps) scores only 27.7 and 34.7 in these measures, respectively. While the best Joint Model substantially outperforms the majority class and random selection baselines, the best Pipeline Model does that only for $F_1$-score but not for accuracy. This indicates the advantage of the Joint Model, which jointly processes the entire set of messages.

For both the Pipeline Model and the Joint Model it is a linear classifier that performs best. Interestingly, for individual messages CNN and the linear classifiers perform similarly well with a slight advantage to CNN (Table~\ref{table:results}), the performance gap is in the main task of location type classification. 
Since the linear classifiers do not consider the relations between the messages, we consider this result as another indication of the challenge the unique structure of our task (modeling weakly-related messages) poses for structure-aware modeling. In fact, as noted in Section~\ref{section:learning}, the results we report for CNN are with random message ordering, as our more informed ordering strategies failed to improve results.

Interestingly, for the best performing models in both the Joint and Pipeline models, LM and POS-based features are not included in the best feature configuration according to the initial experiments. Hence, we do not report test-set results for these features. For the best performing model (logit, Joint Model) the spatio-textual features are also not included in the best feature configuration, as they are not used by the Joint Model (Section~\ref{section:learning}). This emphasizes the importance of shallow textual features for our task and the need for further research into more sophisticated linguistic features as well as the collection of more relevant contextual information regarding the time, location and geography of the messages.

Finally, our choice of hand-crafted features is also supported by an experiment where we trained the linear models with word embedding features (results are not shown in the table). In the Pipeline Model we represented each message by the average of its word embeddings, while in the Joint Model an entire set of messages is represented by the average of its word embeddings. The resulting models were substantially outperformed by the best linear models trained with hand-crafted features. For example, in the Joint Model the logit classifier scores 31.5 in accuracy and 9.1 in $F_1$ (not shown in the table), compared to the respective 52.6 and 48.8 scores of the best performing logit model, as shown in Table~\ref{table:result:1}. The same pattern holds for Naive Base and the Pipeline Model.

\subsection{Ablation Test}
\label{section:ablation}

We compared the different feature families in order to find the most effective one. We measured both accuracy considering each feature family separately and the ablation consisting all the feature families while excluding a single family. An family of features is significant if when excluded the accuracy and $F_1$ scores plummet. The comparison, presented in Table~\ref{table:ablation}, was conducted over a logit classifier for the first step of the two-steps classification, the step of single-tweet classification.

It can be seen that uni-grams and bi-grams are the most significant features both by accuracy and ablation. Textual features are also very important where other features have marginal contribution. This further justifies the Join Model (one-step classification), as discussed in Section~\ref{section:problem}.

\begin{table}[t]
\centering
  \renewcommand{\arraystretch}{1}
  \scalebox{0.9}{
  \begin{tabular}
  {|l|c|c|c|c|}
  \hline
  \multirow{2}{*}{Features} & \multicolumn{2}{c|}{\textbf{Single Feature}} & \multicolumn{2}{c|}{\textbf{All But This Feature}}\\
   & Accuracy & $F_1$ & Accuracy & $F_1$ \\
  \hline
  All features & \multicolumn{2}{c|}{60.36} & \multicolumn{2}{c|}{56.24} \\
  \hdashline
	textual & 34.07 & 18.79 & 60.03 & 55.59\\
  $n$-grams & \textbf{54.96} & \textbf{44.81} & \textbf{38.72} & \textbf{30.09}\\
  POS & 32.74 & 17.14 & 60.34 & 56.23\\
  LMs & 30.19 & 13.85 & 60.36 & 56.25\\
  spatio-temporal & 31.15 & 10.67 & 55.01 & 44.91\\
  \hline 
  \end{tabular}}
  \caption{\label{table:ablation} Ablation test conducted over the first step of two steps model by logit classifier.}
\end{table}
\section{Conclusion}
\label{section:conclusion}

We introduced and studied the task of Geosocial Location Classification, where the type of a geospatial location is discovered based on a set of geottaged social-media messages posted from that place. We analyzed a large number of modeling choices. We compared a Pipeline Model with a Joint Model, hand-crafted versus automatically-learned features and linear (structure-ignorant) and DNN-based (structure-aware) classifiers. Our experiments show that the Joint Model, which jointly processes the entire set of messages associated with the location, using Logit classifier over n-grams provides the best results.

The most powerful features in our experiments were textual and n-gram features, while NLP and spatio-textual features were not included in the best feature configurations. Moreover, linear models that do not explicitly account for inter-message connections outperformed CNN. For DNNs we also observed that exploiting distance and language modeling information for message ordering failed to provide better results than random ordering. 

While our best classifier (Joint Model using Logit classifier over n-grams) significantly outperforms baseline methods (random, majority) and also methods based on deep neural networks, it still has limited accuracy (accuracy and $F_1$ score of approximately 50\%). Several conclusions can be drawn from that. First, our experiments show that there is a real signal in social-media posts regarding the type of the location from which messages were posted, because classification based on message content is much more accurate than random selection of types. Second, our extensive set of experiments and the use of state-of-the-art classification tools show that the classification task we studied is a hard problem.
Third, combining many messages helps tackling the problem of noisy messages. We can see that by comparing the accuracy of classification of sets of messages versus the accuracy of classifying individual messages---the first has a much higher accuracy.

There are different ways to use the results presented in this work for automatic type association. One is by combining the textual classification with other types of classifiers, e.g., information about people that visit each location such as age, frequency of visits in the site, etc. Another way is by adding location type only in cases where there is a high confidence, i.e., given a high score by the Logit classifier. Studying these approaches is future work.






\bibliographystyle{ACM-Reference-Format}
\bibliography{references} 

\end{document}